\documentclass[10pt,journal,compsoc]{IEEEtran}
\usepackage{amssymb}
\ifCLASSINFOpdf
  \usepackage[pdftex]{graphicx}
  \graphicspath{{../pdf/}{../jpeg/}}
  \DeclareGraphicsExtensions{.pdf,.jpeg,.png}
\else

\fi
\usepackage{amsmath}

\usepackage{bbding} 
\usepackage{multirow}
\usepackage{stfloats}
\usepackage{cuted}
\usepackage{capt-of}
\usepackage[noend]{algpseudocode}
\usepackage{algorithmicx,algorithm} 
\usepackage{color}
\usepackage[colorlinks,linkcolor=blue]{hyperref}

\ifCLASSOPTIONcompsoc
  \usepackage[nocompress]{cite}
\else
  \usepackage{cite}
\fi

\ifCLASSINFOpdf
\else
\fi
\begin{document}
%
\title{Point Cloud Instance Segmentation with Semi-supervised Bounding-Box Mining}

\author{Yongbin Liao*, Hongyuan Zhu*, Yanggang Zhang, Chuangguan Ye, Tao Chen$^{\dag}$, and Jiayuan Fan
\thanks{Y. Liao, Y. Zhang, C. Ye, and T. Chen are with the Embedded Deep Learning and Visual Analysis Group, School of Information Science and Technology, Fudan University, Shanghai 200433, China. E-mail: $\left \{ybliao19, ygzhang19, cgye19, eetchen\right \}$@fudan.edu.cn.}
\thanks{H. Zhu is with Agency for Science, Technology and Research, Singapore. E-mail: hongyuanzhu.cn@gmail.com.}
\thanks{Y. Liao and H. Zhu are with equal contribution. T.Chen is the corresponding author. }
\thanks{J. Fan is with the Academy for Engineering and Technology, Fudan University, Shanghai 200433, China. E-mail: jyfan@fudan.edu.cn.}
}

\IEEEtitleabstractindextext{
\vspace{-3pt}
\begin{abstract}
Point cloud instance segmentation has achieved huge progress with the emergence of deep learning. However, these methods are usually data-hungry with expensive and time-consuming dense point cloud annotations. To alleviate the annotation cost, unlabeled or weakly labeled data is still less explored in the task. In this paper, we introduce the first semi-supervised point cloud instance segmentation framework (SPIB) using both labeled and unlabelled bounding boxes as supervision. To be specific, our SPIB architecture involves a two-stage learning procedure. For stage one, a bounding box proposal generation network is trained under a semi-supervised setting with perturbation consistency regularization (SPCR). The regularization works by enforcing an invariance of the bounding box predictions over different perturbations applied to the input point clouds, to provide self-supervision for network learning. For stage two, the bounding box proposals with SPCR are grouped into some subsets, and the instance masks are mined inside each subset with a novel semantic propagation module and a property consistency graph module. Moreover, we introduce a novel occupancy ratio guided refinement module to refine the instance masks. Extensive experiments on the challenging ScanNet v2 dataset demonstrate our method can achieve competitive performance compared with the recent fully-supervised methods. Source codes are available on  \href{https://github.com/Lonepic/SPIB}{https://github.com/Lonepic/SPIB} 
\end{abstract}

\begin{IEEEkeywords}
Semi-supervised learning, weakly-supervised learning, point cloud instance segmentation.
\end{IEEEkeywords}}

\maketitle

\IEEEdisplaynontitleabstractindextext

%
\IEEEpeerreviewmaketitle


\section{Introduction}

\IEEEPARstart{T}{he} recent availability of commercial depth sensors has promoted the emergence of many 3D deep learning analysis methods \cite{Jiang_2020_CVPR,wang2018sgpn,yi2019gspn,pham2019jsis3d,han2020occuseg,lahoud20193d,engelmann20203d}. Point cloud instance segmentation is one of the key research topics in 3D computer vision with applications in VR/AR and robots navigation. It is a challenging task as it needs to simultaneously assign points to each object instance and predict its semantic label. However, most of these methods are fully supervised and heavily rely on dense point-level labeled data, which is rather costly to collect. For example, annotating all points of a ScanNet scene \cite{dai2017scannet}  takes nearly 22.3 minutes \cite{Wei_2020_CVPR}. 

There are a lot of attempts trying to tackle this problem, such as weakly-supervised learning which adopts weaker and less expensive labels and semi-supervised learning which requires only few labeled data. Motivated by these methods \cite{zhou2016learning, selvaraju2017grad,durand2017wildcat,rasmus2015semi,tarvainen2017mean,miyato2018virtual} in 2D images, there are also some attempts on weakly-supervised learning and semi-supervised learning for 3D scene understanding recently. MPRM \cite{Wei_2020_CVPR} proposes a multi-path region mining module for semantic segmentation using scene-level labels. 
Xu \textit{et al}. \cite{xu2020weakly} and CSC \cite{hou2020exploring} explore point cloud semantic segmentation using point labels. SESS \cite{zhao2020sess} takes pure point cloud as input and constructs a semi-supervised architecture leveraging mean-teacher network for object detection.

\begin{figure*}[htbp]
    \centering
    \includegraphics[scale=0.54]{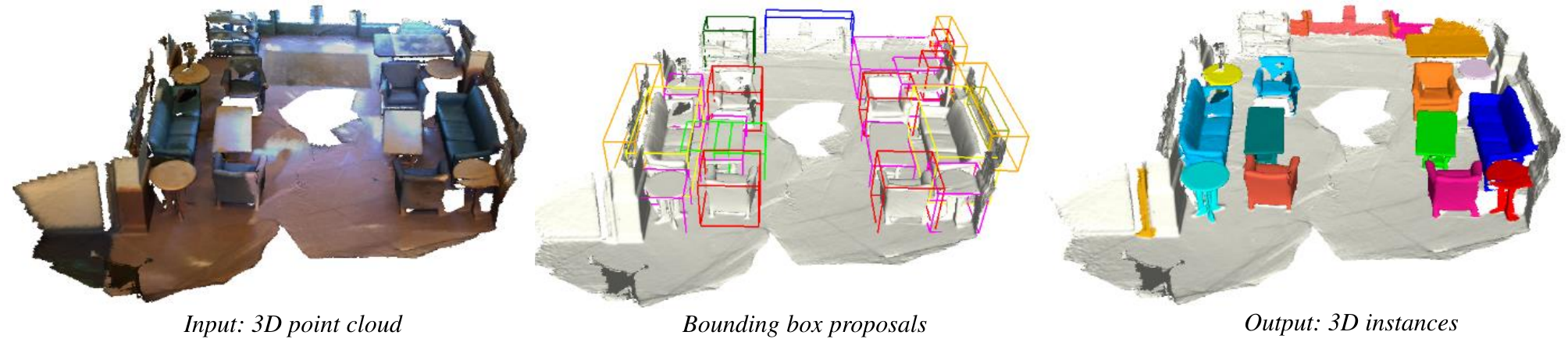}
	\caption{\label{pipeline}Given the input point cloud, we propose the first weakly- and semi-supervised point cloud instance segmentation architecture with bounding box supervision. Bounding box proposals are generated first to group the input point cloud into some point subsets. Afterwards, we predict accurate instance masks with our proposed modules, including semantic propagation component, property consistency graph, and occupancy ratio guided refinement.}
\end{figure*}

In this paper, we introduce a novel semi-supervised framework for point cloud instance segmentation
using 3D bounding boxes as labels, as demonstrated in Fig. \ref{pipeline}. Compared with dense point-level labels, the bounding box label of an instance can be easily annotated by the eight outermost points of that instance, leading to a great reduction of annotation cost and they also provide roughly information of localized instances. Under this setting, we train our SPIB network in two stages: 

At the first stage, we propose a self-supervised perturbation consistency regularization (SPCR) mechanism for semi-supervised bounding box proposal generation. SPCR enforce invariance of the predictions over some perturbations applied to the input point cloud. As a result, unlabeled data could provide extra self-supervision to help on labeled data learning. We propose two consistency terms consisting of both geometric and semantic information for better prediction-invariant constraint under different perturbations. 


At the second stage, given bounding box proposals generated, we predict class scores for each point inside the bounding boxes. The discriminative points are selected when their class scores are above certain thresholds. Then the selected points are propagated to their neighbor points according to their class scores similarities until certain stop conditions are met. 


Moreover, we propose a novel occupancy ratio guided refinement module, using additional important prior constraints provided by counting each category's predicted foreground masks within the bounding boxes, to refine the instance segmentation. 

The main contributions of this paper can be summarized as follows:
\begin{itemize}
\item To the best of our knowledge, this is one of the pioneer works for semi-supervised point cloud instance segmentation, and we are the first to use bounding boxes as supervision in the task.
\item We develop a novel semi-supervised point cloud proposal generation network with perturbation consistency regularization, namely SPCR, to mine the underlying information of unlabeled data for additional self-supervision.
\item We propose a novel semantic propagation component to predict the instances' segmentation. Along with the occupancy ratio guided refinement, instances can be segmented accurately.
\item Results on the challenging ScanNet v2 dataset demonstrate the effectiveness of our method. Specifically, we achieve 33.1\% of mAP@50\% and 54.0\% of mAP@25\% with only  40\% weakly labeled data, which is comparable with the fully supervised counterpart.

\end{itemize}


\section{Related Work}

\textbf{Weakly-supervised Learning on 2D Image:} Recently, weakly supervised segmentation has attracted much attention due to both the labor and time costly data annotation of its fully supervised counterpart. Various types of weak labels such as bounding boxes \cite{dai2015boxsup}, scribbles \cite{lin2016scribblesup,vernaza2017learning}, points \cite{bearman2016s} and image-level classification labels \cite{durand2017wildcat, zhou2018weakly, joon2017exploiting, huang2018weakly, hong2017weakly} have been utilized. For weakly supervised semantic segmentation, most approaches use the image-level supervision based on CAMs \cite{zhou2016learning, selvaraju2017grad}. SEC \cite{kolesnikov2016seed} refines CAMs using seeds. AE-PSL \cite{wei2017object} iteratively erases the discriminative part of CAM to guide the network to learn features from other different regions. AffinityNet \cite{ahn2018learning} learns the pixel-level affinity to generate a transition matrix for CAM adjusting. SEAM \cite{wang2020self} proposes a self-supervised equivalent attention mechanism to provide consistency regularization on predicted CAMs.

For weakly supervised instance segmentation, BoxSup \cite{dai2015boxsup}  iteratively generates region proposals and trains convolutional networks. SDI \cite{khoreva2017simple} combines MCG \cite{pont2016multiscale} and GrabCut \cite{rother2004grabcut} to produce segment proposals, and the generated pseudo ground truth is further used to train their fully supervised instance segmentation network. PRM \cite{zhou2018weakly} detects peak response maps of each class to identify individual instances and utilizes these cues together to predict instance masks. Hsu \textit{et al}. \cite{hsu2019weakly} adopts the Mask R-CNN \cite{he2017mask} framework and formulates the instance segmentation problem as a multiple instance learning (MIL) task. Song \textit{et al}. \cite{song2019box} proposes a box-driven class-wise region masking module and a filling rate loss to remove irrelevant regions of each class for instance mask refinement. IRNet \cite{ahn2019weakly} generates an instance-aware CAM using displace filed and pairwise affinities, and trains the network with inter-pixel relations on the class attention maps. 
 
Motivated by the weakly supervised 2D scene segmentation, we explore weakly supervised point cloud instance segmentation for the first time. In particular, we use bounding boxes as weak supervision. Given the detected region proposal, we adopt seed region growing inside each point subset using our proposed novel semantic propagation module and property consistency graph module. Further, we propose a novel occupancy ratio guided refinement module as an important prior to improve predicted segmentation.

\textbf{Semi-supervised Learning on 2D Image:} Many efforts have been made to adopt semi-supervised methods in deep learning by virtue of its aim to mine potential knowledge of unlabeled data. For example, The $\Gamma$ model \cite{tarvainen2017mean} proposes a ladder network and simultaneously minimizes the loss function of supervised and unsupervised branches by back propagation. The $\Pi $-model \cite{laine2016temporal} encourages a consistency over two different perturbations applied to one input image. Mean-teacher \cite{tarvainen2017mean} enforces similar predictions of student network and teacher network whose weights are transferred from the student. VAT \cite{miyato2018virtual} improves the prediction by approximating the perturbations which influence model’s results the most. Temporal model \cite{laine2016temporal} introduces a self-ensembling method to form a consistent prediction of the unlabeled data using the outputs on different epochs. FixMatch \cite{sohn2020fixmatch} demonstrates the importance of strong and varied perturbations and enforces an implicit entropy minimization for high-confidence predictions on unlabeled data. 

Considering the promising potential of semi-supervised learning and inspired by the recent semi-supervised methods, we try to leverage the setting of semi-supervised learning for proposal generation and propose a perturbation consistency regularization mechanism. The perturbation consistency regularization mechanism is, to keep the predictions over different perturbations applied to the input point cloud as similar as possible. We first propose various perturbation schemes for the input point cloud to mine the potential of unlabeled data to full advantage. Moreover, two consistency terms of both geometric and semantic are introduced for better perturbation consistency regularization.

\textbf{Point cloud Instance Segmentation:} Generally, there are two major categories of 3D instance segmentation methods. The first is proposal-free approaches, where point-level semantic labels are predicted followed by grouping points into object instances utilizing point embeddings. SGPN \cite{wang2018sgpn} groups points based on similarity matrix measured by the difference of feature vectors for all point pairs. JSIS3D \cite{pham2019jsis3d} develops a multi-task point-wise network and a multi-value Conditional Random Field (CRF) component to jointly predict semantic and instance labels. ASIS \cite{wang2019associatively} and BAN \cite{wu2020bi} utilize discriminative loss to pull embeddings of the same instance towards their mean embeddings while pushing mean embeddings of different instances apart. MTML \cite{lahoud20193d} proposes a multi-task metric learning strategy where points with the same instance label are grouped closer to each other, while clusters with different instance labels are more separated from each other. OccuSeg \cite{han2020occuseg} introduces an occupancy-aware scheme, which produces both occupancy signal and embedding representations for graph-cut instance clustering. Recently, \cite{hou2020exploring} proposes an unsupervised pre-training model, which could be fine-tuned on the downstream tasks using point-level labels. However, it needs to actively query an annotator to label some data points that can help for subsequent training, which increases the cycle time of learning. PE \cite{zhang2021point} proposes a probabilistic embedding space for point cloud embedding, and followed a novel loss function which benefits both the semantic segmentation and the clustering.

The other category is proposal-based approaches, which firstly generate region proposals and further predict foreground mask inside each proposal. 3D-SIS \cite{hou20193d} jointly takes multi-view images and point cloud as inputs, followed by dense 3D convolution for object bounding boxes generation and corresponding instance masks prediction. GSPN \cite{yi2019gspn} proposes a generative shape proposal network and identifies instances according to object proposals. 3D-BoNet \cite{yang2019learning} directly predicts bounding boxes of objects and point masks of instances simultaneously. 3D-MPA \cite{engelmann20203d} applies object center voting for proposal generation and gets the final instances by multiple proposal aggregation. PointGroup \cite{Jiang_2020_CVPR} proposes a clustering algorithm and adopts it on the original point set as well as offset-shifted point coordinate set to generate some instance candidates. GICN \cite{liu2020learning} approximates the distributions of instance centers as gaussian center heatmaps and selects some center candidates for bounding boxes and instance masks prediction. SSTNet\cite{liang2021instance} introduces an end-to-end solution of Semantic Superpoint Tree Network for proposing object instances from scene points. HAIS \cite{chen2021hierarchical} makes full use of spatial relation of points and point sets and introduces the hierarchical aggregation to progressively generate instance proposals. 

Likewise, our system adopts proposal-based methods as well since they have better performance and the bounding box proposals provide many localized cues of objects, which is of vital importance to instance segmentation. However, our method learns dense point cloud instance segmentation using a few bounding boxes as weak supervision, which significantly reduces the labeling efforts. 
 
\textbf{Weakly/Semi- supervised Point Cloud Segmentation:} There are some attempts that have been made for weakly/semi- supervised point cloud segmentation. MPRM \cite{Wei_2020_CVPR} proposes a multi-path region mining module to generate pseudo point-level labels, from a classification network trained with scene-level and subcloud-level labels. After that, a point cloud semantic segmentation network is then trained with these point-level pseudo labels in a fully supervised manner. SegGroup \cite{tao2020seggroup} proposes a new form of segmentation-evel weak labels obtained from over segmentation by annotating one point for each instance. Further, SegGroup network is proposed to generate pseudo point-level labels so that existing methods can directly consume the pseudo labels for training. Xu \textit{et al}. \cite{xu2020weakly} propose a weakly supervised point cloud semantic segmentation approach, which requires only a tiny fraction of points to be labeled in the training stage, and reach even better performance compared with the fully-supervised methods. PointContrast \cite{xie2020pointcontrast} proposes an unsupervised pre-training method and selects a suite of diverse datasets and tasks to measure effectiveness, including semantic segmentation and instance segmentation. CSC \cite{hou2020exploring} expands PointContrast by making use of both point-level correspondences and spatial contexts in a scene, and achieves promising performance using only 0.1$\%$ point labels.  Mei \textit{et al}. \cite{mei2019semantic} introduce a semi-supervised point cloud semantic segmentation network, by training a CNN-based classifier considering both supervised samples with manually labeled object classes and pairwise constraints. SSPC-Net \cite{cheng2021sspc} trains a semi-supervised semantic segmentation network by inferring the labels of unlabeled points from the few annotated points, and builds superpoint graphs to mine the long-range dependencies in point clouds for dynamic label propagation.

However, the performance of these methods is still far behind the recent fully-supervised methods. All of these methods are merely focused on weakly-supervised or semi-supervised point cloud semantic segmentation solely. In this paper, we explore a better trade-off between annotation cost and performance, and propose a weakly- and semi-supervised point cloud instance segmentation network with box-level labels.

\begin{figure*}[htbp]
    \centering
    \includegraphics[scale=0.42]{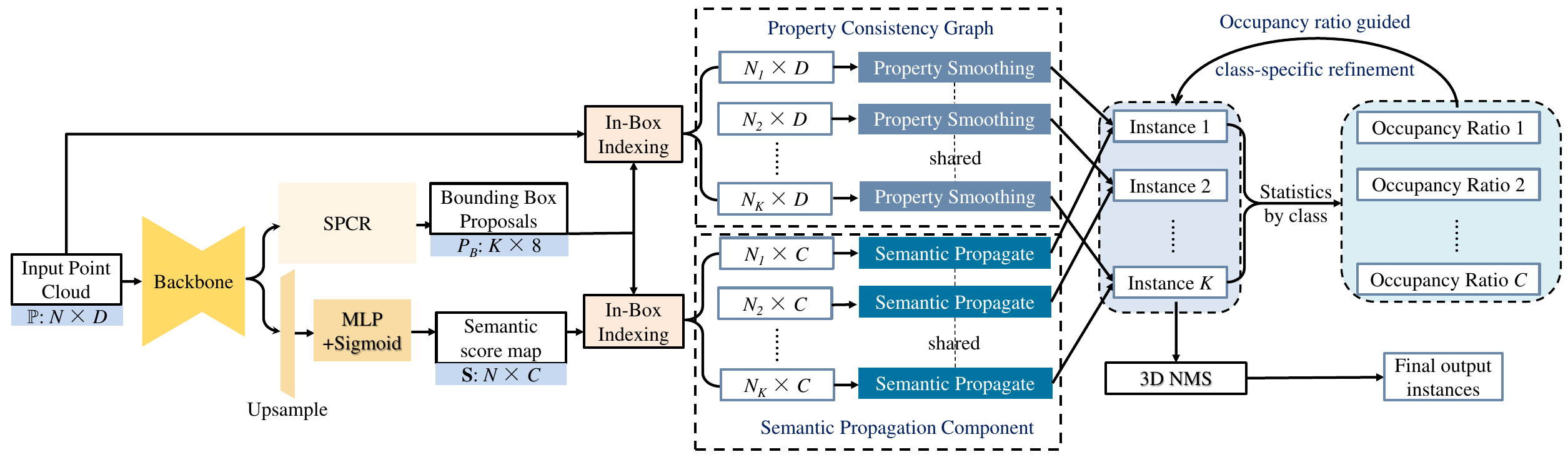}
	\caption{\label{SPIB}Illustration of the SPIB architecture. From an input point cloud $\mathbb{P}$, we first generate bounding box proposals through the proposal generation network SPCR. We then conduct an in-box indexing to select a fixed \textit{K} number of point subsets both in the original input point cloud and the learned semantic score map. Within these point subsets, we predict instance masks through our proposed semantic propagation component and property consistency graph, thus produce \textit{K} instance candidates. Afterwards, category-based statistics of occupancy ratio are computed for \textit{C} semantic classes, followed by class-specific occupancy ratio guided refinement for each instance candidate. Finally, we remove duplicate instances by 3D Non-Maximum Suppression (NMS) and output the final 3D instances.}
\end{figure*}

\begin{figure*}[htbp]
    \centering
    \includegraphics[scale=0.53]{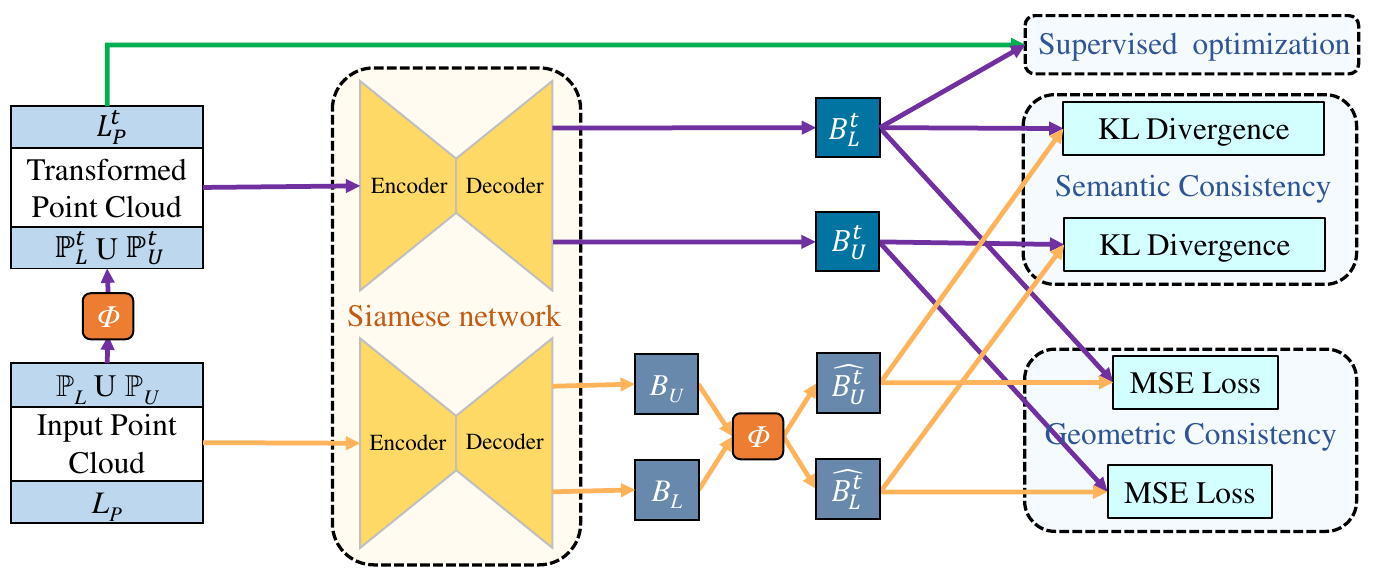}
	\caption{\label{SPCR} The network architecture of our proposed SPCR. Given the original input point cloud (either labeled or unlabeled) and their transformed point cloud, we first generate bounding box proposals through the Siamese network. The predictions of the original point cloud are then transformed with the same perturbation $\mathit{\Phi}$. Finally, these bounding box proposals are optimized by the supervised loss and our proposed perturbation consistency regularization mechanism.}
\end{figure*}

\section{Our Method}
\label{sec:method}
We propose a semi-supervised point cloud instance segmentation framework (SPIB) with bounding boxes as labels, as shown in Fig.\ref{SPIB}. Specifically, our architecture contains two main stages: we firstly leverage a network named semi-supervised proposal generation with perturbation consistency regularization (SPCR), to generate bounding box proposals, which can be further used to get coarse point cloud instance segmentation by treating all the points within the bounding box as an instance. After that, the coarse point cloud instance segmentation from SPCR is further refined by three proposed modules using bounding boxes as supervision: semantic propagation module, property consistency graph module, and occupancy ratio guided refinement module. The semantic propagation module propagates seeded regions through the semantic similarity between the seed points and their neighbors. The property consistency graph module encourages the points with similar properties to be grouped as the same instance. The occupancy ratio guided refinement module refines each instance proposal according to the mean occupancy ratio of its corresponding class. With these two steps, a semi-supervised point cloud instance segmentation architecture using bounding boxes as supervision (SPIB) is accomplished. 

\subsection{Semi-supervised Bounding Box Proposal Generation with Pertubation Consistency Regularization (SPCR)}
\label{sec3.1}
\subsubsection{SPCR Architecture}
The overall architecture of our proposed SPCR is depicted in Fig.~\ref{SPCR}. We introduce the implementation of semi-supervised bounding box proposal generation by a weight-shared siamese network which is composed of the VoteNet \cite{qi2019deep}. To be specific, we take a mixture of labeled and unlabeled point clouds as our input denoted as $\left \{  \mathbb{P}_{L} \cup \mathbb{P}_{U}\right \}$, where $\mathbb{P}_{L}$ and $\mathbb{P}_{U}$ represent the labeled and unlabeled point cloud respectively. To extract additional training signal for self-supervision, we conduct a perturbation $\mathit{\Phi}$ on the original point cloud to obtain transformed point cloud denoted as $\left \{ \mathbb{P}_{L}^{t} \cup \mathbb{P}_{U}^{t} \right \}$. After that, the original point cloud and transformed point cloud are passed to the siamese network simultaneously for proposal generation. The output bounding box proposals are represented by $\left \{ B_{L},B_{U} \right \}$ and $\left \{ B_{L}^{t},B_{U}^{t} \right \}$ respectively. $\left \{ B_{L},B_{U} \right \}$ are further transformed to $\left \{ \hat{B_{L}},\hat{B_{U}} \right \}$ by the same perturbation conducted on the original point cloud. For bounding box proposals $B_{L}^{t}$, we optimize them with transformed ground truths $L_{P}^{t}$ through supervised loss defined in VoteNet. For other predictions, we propose a perturbation consistency regularization to constrain the pair-wise proposals to be consistent both in semantic categories and geometric properties. Details of the perturbation scheme and perturbation consistency regularization mechanism will be described in the next two subsections.

\subsubsection{Perturbation Scheme}
An important factor for consistency regularization is the perturbations applied to the input point cloud. We propose three types of perturbations consisting of random jittering, flipping and rotation to prevent the labeled data from overfitting and leverage the unlabeled data for self-supervised learning. At first, we initialize the perturbation matrix \textbf{M} of $\mathit{\Phi}$ with an identity matrix of shape 3 $\times$ 3. For jittering, the perturbation matrix will be added with a random matrix of the same shape as \textbf{M}. For flipping, we update \textbf{M} by multiplying its first element with a random variable selected from $ \left \{ 1,-1 \right \}$, where -1 means flipping along y-axis and 1 means no flipping. For rotation, we firstly generate a rotation angel formulated as $\theta = 2\pi \delta$ where $\delta$ is a random variable, and further define the corresponding rotation matrix as 
\begin{equation}
R(\theta )=\begin{bmatrix}
cos(\theta)  & sin(\theta) & 0\\ 
-sin(\theta)&  cos(\theta)& 0\\ 
0 & 0 & 1
\end{bmatrix}
\end{equation}
After that, we will adjust \textbf{M} by multiplying it with rotation matrix $R(\theta )$ and thus get the final perturbation matrix. 

With the established perturbation matrix \textbf{M} computed via the above random perturbation scheme, a training batch with a mixture of labeled and unlabeled samples will be transformed by multiplying it with \textbf{M}. In particular, the labels $\mathit{ L_{P}}$ of labeled samples $\mathbb{P}_{L}$ are also transformed by the same perturbation before supervised optimization. To ensure the validity of  consistency regularization, the instance predictions $\left \{ B_{L},B_{U} \right \}$ of original inputs $\left \{  \mathbb{P}_{L} \cup \mathbb{P}_{U}\right \}$ are transformed as well.

\subsubsection{Consistency Regularization}
 As stated above, in order to provide additional self-supervision for network training, we rely on enforcing a consistency regularization between the predictions under different perturbations. It is obvious that the consistency of predictions should include both semantic category level and geometric property level. To this end, we propose two terms of consistency regularization for semi-supervised point cloud instance segmentation and define the consistency regularization loss as 
\begin{equation}
L_{CR}= \lambda_1 L_{semantic}+\lambda_2 L_{geometric}
\end{equation}
where the semantic term $L_{semantic}$ enables consistency regularization by enforcing the similar predictions of semantic categories, and the geometric term imposes the structural constraint on bounding box proposals for maintaining consistent geometric properties. The $\lambda_1$ and $\lambda_2$ are the weights to control the importance for each term.
 
\textbf{[a] Semantic Consistency Term}: For predictions $\left \{ B_{L}^{t},B_{U}^{t} \right \}$ and $\left \{ \hat{B_{L}^{t}}, \hat{B_{U}^{t}} \right \}$ of the original and transformed input point clouds, let $\left \{ S_{L}^{t},S_{U}^{t} \right \}$ and $\left \{ \hat{S_{L}^{t}},\hat{S_{U}^{t}} \right \}$ represent the semantic probabilities of these bounding box proposals respectively. We define the semantic consistency regularization term as the KL-divergence:  
\begin{equation}
L_{semantic}=\frac{\sum D_{KL}(s_{L}^{t}\parallel\hat{s_{L}^{t}} )+ \sum D_{KL}(s_{U}^{t}\parallel\hat{s_{U}^{t}} )}{\left | \hat{S_{L}^{t}} \right |+\left | \hat{S_{U}^{t}} \right |}
\end{equation}

\textbf{[b] Geometric Consistency Term}: Using the semantic term alone can only constrain the predicted semantic class of bounding box proposals, while ignore another import constraint of geometric information. Similarly, we denote the center of bounding box proposals as $\left \{ O_{L}^{t},O_{U}^{t} \right \}$ and $\left \{ \hat{O_{L}^{t}},\hat{O_{U}^{t}} \right \}$ respectively. The geometric consistency regularization is then formulated as 
\begin{equation}
L_{geometric}=\frac{\sum(o_{L}^{t}-\hat{o_{L}^{t}} )+ \sum (o_{U}^{t}-\hat{o_{U}^{t}} )}{\left | \hat{O_{L}^{t}} \right |+\left | \hat{O_{U}^{t}} \right |}
\end{equation}


\subsection{Semi-supervised Point Cloud Instance Segmentation with Bounding Box Labels (SPIB)}
As demonstrated in Fig.~\ref{SPIB}, the input of our SPIB architecture is a point cloud $\mathbb{P}$ of size $N \times D$, consisting of  \textit{N} points and \textit{D}-dimensional input features (e.g. coordinates, colors, and normals). Following the procedure of \cite{qi2019deep}, we construct our backbone network based on PointNet++ using four set abstraction layers and two feature propagation (upsampling) layers. We then feed the point-wise features extracted by this backbone network into two branches, one for semantic label prediction and the other for proposal generation. 

For proposal generation, we generate a fixed number \textit{K} of bounding box proposals through our proposed SPCR and we denote the proposals as $P_B=\left \{ P_{B_{i}} \right \}_{i=1}^{K}\in \mathbb{R}^{K\times 8\times 3}$. Each proposal is parameterized by its 8 outermost vertices and its corresponding GT semantic class is $C_{i}$. 

For semantic label prediction, two feature propagation layers with skip link to its corresponding set abstraction layers are added after the backbone network, to upsample the feature map to the same scale of the original input point cloud. An MLP layer followed by \textit{sigmoid} function is applied to produce semantic scores for the input \textit{N} points over \textit{C} semantic categories. We denote the predicted semantic score map as $\mathbf{S} \in \mathbb{R}^{N\times C}$.

After the proposals are predicted, initial seeded regions are generated by selecting points inside each bounding box proposal, and thus form \textit{K} number of point subsets denoted as $\mathbf{P}=\left \{ P_{1(N_{1})},P_{2(N_{2})},...,P_{i(N_{i})},...,P_{K(N_{K})} \right \}$, where $P_{i}$ is the points inside the \textit{i}-th bounding box proposal, \textit{K} is the fixed number of proposals and $N_{i}$ is used to represent the number of points in $P_{i}$. Afterwards, we start to mine the accurate instance masks based on the original input point cloud and the learned semantic score map. Details of them will be described in the next three sub-sections with pseudo code of the learning process in Algorithm 1.

\subsubsection{Semantic Propagation Component}
\label{sec. 3.2}
Given the predicted semantic score map $\mathbf{S} \in \mathbb{R}^{N\times C}$ and the separated \textit{K} number of point subsets $\mathbf{P}=\left \{ P_{1(N_{1})},...,P_{i(N_{i})},...,P_{K(N_{K})} \right \}$, the goal of this stage is to extract the foreground region within each point subset.

To start, for each point subset, we gather the point semantic scores from \textbf{S} and thus form $\mathbf{S}_{P_{i}} = \left \{ S_{I(P_{i},1)},...,S_{I(P_{i},N_{i})} \right \}\in \mathbb{R}^{N_{i}\times C}$, where \textit{I} maps the point index in $P_{i}$ to corresponding point index in $\mathbb{P}$. And then, we introduce a soft label to supervise the process of semantic prediction for each point as follows:
\begin{equation}
l_{P_{ij}}\!=\!\left\{\begin{matrix}
0, & argmax(\mathbf{S}_{P_{ij}})\!\neq\!  C_{i}&and&{\mathbf{S}_{P_{ij}}}'\!<\!  \delta _{l} \\ 
1, & argmax(\mathbf{S}_{P_{ij}})\!=\! C_{i}&and&{\mathbf{S}_{P_{ij}}}'\!>\! \delta _{h}\\ 
\frac{{\mathbf{S}_{P_{ij}}}'-\delta _{l}}{\delta _{h}-\delta _{l}}, & otherwise&
\end{matrix}\right. \label{equation5}
\end{equation}
where $j\in\left \{ 1,...,N_{i} \right \}$, $C_{i}$ represents the predicted category of the bounding box proposal, $\delta _{h}$ and $\delta _{l}$ are empirically set to 0.3 and 0.7 respectively in our implementation. The ${\mathbf{S}_{P_{ij}}}'$ is the normalized semantic scores formulated as
\begin{equation}
{\mathbf{S}_{P_{i}}}' = \frac{\mathbf{S}_{P_{i}}[C_{i}]}{max(\mathbf{S}_{P_{i}}[C_{i}])}
\end{equation}
where $\mathbf{S}_{P_{i}}[C_{i}]$ indicates the semantic scores on channel $C_{i}$ of point subset $P_{i}$. Therefore, the points assigned with label 1 as foreground points form the seed region of one instance candidate.

\begin{figure}[]
    \centering
    \includegraphics[scale=1.25]{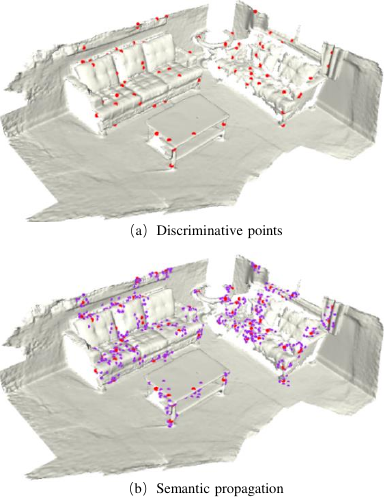}
	\caption{\label{sp} Illustration of semantic propagation. Given the predicted discriminative points (red) of each object, we propagate its seed region by adding the highly similar points (purple) within the neighborhood balls of discriminative points. Points are enlarged for better visibility.}
\end{figure}

However, seed regions extract merely the discriminative parts of an instance rather than the whole instance, so we conduct semantic propagation to address this issue as shown in Fig.~\ref{sp}. For each discriminative point inside one seed region, we find all neighbor points that are within a radius to it and measure the semantic similarity between them, to force the neighbor points with high semantic similarity to have close semantic scores to their corresponding discriminative point. The semantic similarity is defined as 
\begin{equation}
\begin{split}
ss_{jk}=exp(-\frac{\left \| {\mathbf{S}_{P_{ij}}}'-{\mathbf{S}_{P_{ik}}}' \right \|^{2}}{\sigma })m_{jk} \\
\forall j\in\left \{ 1,...,N_{i} \right \},k\in\left \{ 1,...,{N_{i}}' \right \} 
\end{split} \label{equation7}
\end{equation}
where ${N_{i}}'$ represents the number of discriminative points in \textit{i}-th point subset. The $m_{jk}$ is a binary mask, $m_{jk}=1$ if $\left \| x_{j}-x_{k} \right \|< r$ and $m_{jk}=0$ otherwise, \textit{r} is the radius of the neighborhood ball, $x_{k}$ and $x_{j}$ are coordinates of discriminative point \textit{k} and neighbor point \textit{j} respectively. 

During training, we include the binary cross-entropy loss which is used to supervise the region growing process:
\begin{equation}
\begin{split}
L_{seed} = -\frac{1}{K}\frac{1}{N_{i}}\sum_{i=1}^{K}\sum_{j=1}^{N_{i}}[l_{P_{ij}}log({\mathbf{S}_{P_{ij}}}') \\+ (1-l_{P_{ij}})log(1-{\mathbf{S}_{P_{ij}}}') ] \label{equation8}
\end{split}
\end{equation}
and we further define the semantic propagation loss as
\begin{small}
\begin{equation}
 L_{SP}=\frac{1}{K}\frac{1}{{N_{i}}'}\frac{1}{N_{i}}\sum_{i=1}^{K}\sum_{k=1}^{{N_{i}}'}\sum_{j=1}^{N_{i}}ss_{jk}\left \| {\boldsymbol{S}_{P_{ij}}}' - {\boldsymbol{S}_{P_{ik}}}' \right \|^{2} \label{equation9}
\end{equation}
\end{small}to encourage nearby points with similar semantics to take similar labels.

\subsubsection{Property Consistency Graph}

We observed that the same instance always contains points with similar properties such as coordinates, colors and normals. To this end, we construct a property consistency graph \textit{G} to promote the effect of instance mask prediction within each point subset $P_{i(N_{i})}$. Each point in $P_{i(N_{i})}$ is represented by its internal property, i.e. located coordinate $l_{a} = [x_{a},y_{a},z_{a}]$, color $c_{a} = [r_{a},g_{a},b_{a}]$ and normal $n_{a} = [nx_{a},ny_{a},nz_{a}]$. The weight matrix of this graph is defined as 
\begin{small}
\begin{equation}
\hspace{-1mm}
w_{ab} \!=\!\\
\left\{\begin{matrix}
exp(-\frac{{\left \| l_{a}\!-\!l_{b} \right \|}^2}{\theta _{1}}-\frac{{\left \| c_{a}\!-\!c_{b} \right \|}^2}{\theta _{2}}-\frac{{\left \| n_{a}\!-\!n_{b} \right \|}^2}{\theta _{3}}), & b\!\in\! knn(P_{ia})\\ 
 0, & b\!\notin\!  knn(P_{ia})
\end{matrix}\right. \label{equation10}
\end{equation}
\end{small} where $a,b\in\left \{ 1,...,N_{i} \right \}$, $knn(P_{ia})$ is the \textit{k} nearest neighbor points of point \textit{a}, which is dependent on the property disparity matrix formulated as 
\begin{equation}
D_{ab} = \left \| l_{a}-l_{b} \right \|^{2} + \left \| c_{a}-c_{b} \right \|^{2}+\left \| n_{a}-n_{b} \right \|^{2}
\end{equation} consisting of pairwise distance on three internal properties: coordinates, colors and normals.

With the constructed Graph \textit{G}, we encourage the prediction $\mathbf{S}_{P_{ia}}$ and $\mathbf{S}_{P_{ib}}$ to be close to each other, if  the weight $w_{ab}$ indicates high property correlation between vertex \textit{a} and \textit{b}. Thus, the loss of this section is given by 
\begin{equation}
L_{PC} = \frac{1}{K}\frac{1}{w_{N}}\sum_{i=1}^{K}\sum_{a}^{}\sum_{b}^{}w_{ab}\left \| \mathbf{S}_{P_{ia}}-\mathbf{S}_{P_{ib}} \right \|^{2} \label{equation12}
\end{equation}
where $w_{N}$ is the sum number of $w_{ab}$ which is not zero.

\begin{figure}[]
    \centering
    \includegraphics[scale=1]{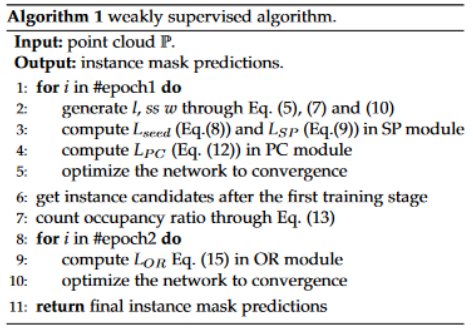}
\end{figure}

\subsubsection{Occupancy Ratio Guided Refinement}
With the semantic propagation module and property consistency graph, the point clouds are coarsely segmented. However, to produce more accurate segmentation, one should focus more on the foreground objects and eliminate the background regions. 

Note that the bounding boxes contain much semantic and object information. A straightforward idea is to learn a global mask to help remove the backgrounds in the scenes. However, such ground-truth mask information is not available in the weakly supervised setting. 

It is well known that the score map in a well-trained model has different response values, indicating the conﬁdence of prediction. A natural idea is to select the locations with the largest confident scores for backward learning, whereas ignore the less conﬁdent ones. However, it is difficult to determine a foreground/backgorund threshold for different classes.

\begin{figure}[]
    \centering
    \includegraphics[scale=0.295]{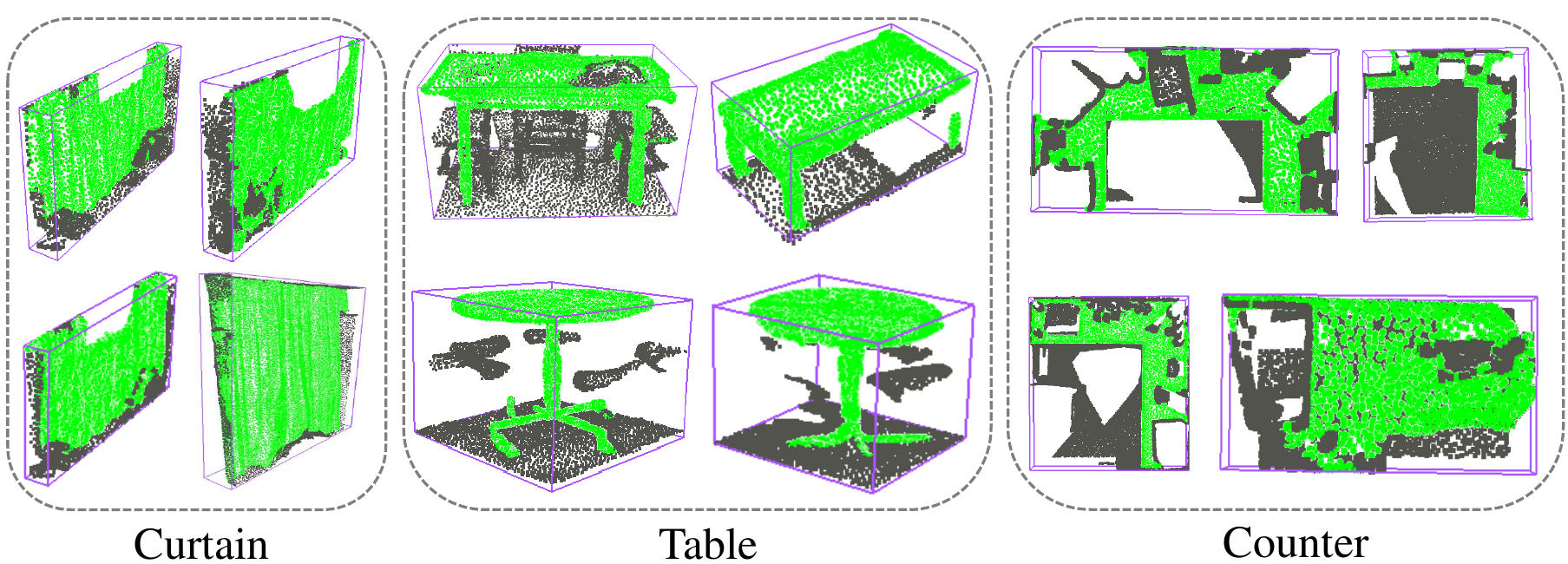}
	\caption{\label{occupancy}Examples of different classes' instance predictions. The predicted instance points are denoted in green and non-instance points within the bounding box are shown in gray. It is obvious that the occupancy ratios of the same class are similar and different classes vary from each other.}
\end{figure}

On the other hand, different classes usually have different shapes. As shown in Fig.~\ref{occupancy}, curtain has $84\%$ foreground points within its predicted bounding box while counter only occupies $54\%$ points in the ScanNet v2 dataset. This inspires us to calculate the mean occupancy rates of each class's predictions as a strong shape prior, since different classes' occupancy ratios are usually different.  

Specifically, we calculate the ratio of predicted instance points to all the points within their predicted bounding boxes for each semantic class, to form the occupancy ratio of this class. For semantic class $C_{i}$, we denote its occupancy ratio as 
\begin{equation}
O_{C_{i}} = \frac{1}{M_{C_{i}}}\sum_{m=1}^{M_{C_{i}}}\frac{P_{ins}(m)}{P_{box}(m)} \label{equation13}
\end{equation}where $M_{C_{i}}$ represents the total number of instances in class $C_{i}$, $P_{ins}(m)$ and $P_{box}(m)$ are the number of points on \textit{m}-th instance proposal and the number of points in corresponding bounding box respectively. 

As a result, we can determine how much percent of points could be regarded as positive points on each instance proposal. Thus, the soft label used in Sec.\ref{sec. 3.2} should be changed to a binary label as 
\begin{equation}
l_{P_{ij}} = \left\{\begin{matrix}
1 &  j\in \text{top} (O_{C_{i}} \times N_{i})\\ 
0 & \text{otherwise}\\ 
\end{matrix}\right.
\end{equation}where the metrics of function \textit{top()} is the predicted semantic scores of points on point subset $P_{i}$. And the loss is also the binary cross-entropy loss given by
\begin{equation}
\begin{split}
L_{OR} = -\frac{1}{K}\frac{1}{N_{i}}\sum_{i=1}^{K}\sum_{j=1}^{N_{i}} [l_{P_{ij}}log({\mathbf{S}_{P_{ij}}}') \\
+ (1-l_{P_{ij}})log(1-{\mathbf{S}_{P_{ij}}}')] \label{equation15}
\end{split}
\end{equation}

\section{Experiments}
\subsection{Dataset and Evaluation Criteria}
\textbf{Dataset.} We evaluate our semi-supervised point cloud instance segmentation method on the challenging ScanNet v2 \cite{dai2017scannet} dataset. The dataset contains 1613 scans, which are officially split into 1201 training samples, 312 validation samples, and 100 testing samples. Each scene of training and validation sets is well annotated with axis aligned bounding boxes and point-level semantic-instance labels.

\textbf{Evaluation Criteria.} Following the common experimental protocol for point cloud instance segmentation, we adopt the same 18 object classes for evaluation as reported in \cite{Jiang_2020_CVPR}. We use the mean average precision at Intersection-over-Union (IoU) threshold 0.25 (mAP@25$\%$), as well as threshold 0.5 (mAP@50$\%$) as our evaluation criteria as defined in the ScanNet benchmark. 

\subsection{Implementation Details}

\textbf{Network:} The input to our network are batches of point clouds with \textit{N} points randomly sub-sampled from the original scan scenes, where \textit{N} is set as 40000. We adopt VoteNet \cite{qi2019deep} as the structure of the siamese network in SPCR. The \textit{K} number of proposals is set as 256. For semantic propagation component, we set \textit{r} = 0.03m and $\sigma$ = 1000. For property consistency graph, variance $\theta_{1},\theta_{2},\theta_{3}$ are set as 1000 and \textit{k} is chosen as 10 empirically. 

\textbf{Learning:} We train the network with semantic propagation component and property consistency graph module for 180 epochs (\#epoch1) with batch size 8, and fine-tune them with occupancy ratio guided refinement for 50 more epochs (\#epoch2). We take Adam as the optimizer with an initial learning rate of 0.001, which is decreased by 10 at the 80-th epoch and then decreased by 10 at the 120-th epoch.

\textbf{Inference:} During inference, our network takes the point clouds of entire scenes as inputs and produces instance candidates. These instance candidates are then fed into 3D NMS module to reduce duplicates with an IoU threshold of 0.25. Different from \cite{qi2019deep}, the IoU here is computed according to instances rather than bounding boxes.

\begin{table*}[]
\caption{\label{semi_weak} Comparison of different label ratios with VoteNet and SPCR on ScanNet v2 val set. Absolute improvements between our SPIB and VoteNet are reported in terms of mAP@50$\%$ and mAP@25$\%$ respectively.}
\renewcommand\arraystretch{1.5}
\setlength{\tabcolsep}{4mm}{
\scalebox{1.0}{
\begin{tabular}{c||c|cccccccccc}
\hline
\hline
Method                        & Metric   & 10\% & 20\% & 30\% & 40\% & 50\% & 60\% & 70\% & 80\% & 90\% & 100\% \\ \hline
\multirow{2}{*}{VoteNet \cite{qi2019deep}}      & mAP@50\% & 9.6  & 15.2 & 22.1 & 24.3 & 26.4 & 28.5 & 29.2 & 30.5 & 32.1 & 33.4  \\
                              & mAP@25\% & 26.3 & 36.9 & 43.8 & 46.1 & 49.2 & 52.3 & 51.1 & 53.5 & 54.2 & 54.8  \\ \hline
\multirow{2}{*}{SPCR}         & mAP@50\% & 13.2 & 20.4 & 22.9 & 30.3 & 29.2 & 27.2 & 30.5 & 33.0 & 33.2 & 35.2  \\
                              & mAP@25\% & 31.4 & 40.3 & 45.8 & 50.4 & 51.7 & 51.2 & 50.9 & 55.6 & 55.3 & 57.9  \\ \hline
\multirow{2}{*}{SPIB}         & mAP@50\% & 19.0 & 26.1 & 29.1 & 33.1 & 31.9 & 32.2 & 32.8 & 36.9 & 36.5 & 38.6  \\
                              & mAP@25\% & 38.8 & 45.4 & 51.4 & 54.0 & 54.5 & 55.4 & 54.9 & 58.4 & 58.6 & 61.4  \\ \hline
\multirow{2}{*}{Improvements} & mAP@50\% & 9.4  & \textbf{10.9} & 7.0  & \textbf{8.8}  & \textbf{5.5}  & \textbf{3.7}  & 3.6  & \textbf{6.4}  & 4.4  & 5.2   \\
                              & mAP@25\% & \textbf{12.5} & 8.5  &\textbf{ 7.6}  & 7.9  & 5.3  & 3.1  & \textbf{3.8}  & 4.9  & \textbf{4.4}  & \textbf{6.6}   \\ \hline \hline
\end{tabular}}}
\end{table*}

\begin{table*}[]
\caption{\label{full_25}Comparison of fully supervised methods on ScanNet v2. The mAP@25$\%$ score is reported on validation set. SP represents semantic propagation component, PC represents property consistency graph, and OR represents occupancy ratio guided refinement.}
\renewcommand\arraystretch{1.5}
\setlength{\tabcolsep}{0.9mm}{
\scalebox{1.0}{
\begin{tabular}{c||c|c|cccccccccccccccccc}
\hline
\hline
Mode                  & Method       & mAP@25\% & cab  & bed  & chair & sofa & tabl & door & wind & bkshf & pic  & cntr & desk & curt & fridg & showr & toil  & sink & bath & ofurn \\ \hline
\multirow{7}{*}{Full} & SegCluster \cite{hou20193d}   & 13.4     & 11.8 & 13.5 & 18.9  & 14.6 & 13.8 & 11.1 & 11.5 & 11.7  & 0.0  & 13.7 & 12.2 & 12.4 & 11.2  & 18.0  & 19.5  & 18.9 & 16.4 & 12.2  \\
                      & MRCNN \cite{he2017mask}       & 17.1     & 15.7 & 15.4 & 16.4  & 16.2 & 14.9 & 12.5 & 11.6 & 11.8  & 19.5 & 13.7 & 14.4 & 14.7 & 21.6  & 18.5  & 25.0  & 24.5 & 24.5 & 16.9  \\
                      & SGPN \cite{wang2018sgpn}        & 22.2     & 20.7 & 31.5 & 31.6  & 40.6 & 31.9 & 16.6 & 15.3 & 13.6  & 0.0  & 17.4 & 14.1 & 22.2 & 0.0   & 0.0   & 72.9  & 52.4 & 0.0  & 18.6  \\
                      & 3D-SIS \cite{hou20193d}      & 35.7     & 32.0 & 66.3 & 65.3  & 56.4 & 29.4 & 26.7 & 10.1 & 16.9  & 0.0  & 22.1 & 35.1 & 22.6 & 28.6  & 37.2  & 74.9  & 39.6 & 57.6 & 21.1  \\
                      & MTML \cite{lahoud20193d}        & 55.4     & 34.6 & 80.6 & 87.7  & 80.3 & 67.4 & 45.8 & 47.2 & 45.3  & 19.8 & 9.7  & 49.9 & 54.2 & 44.1  & 74.9  & 98.0  & 44.5 & 79.4 & 33.5  \\
                      & PointGroup \cite{Jiang_2020_CVPR}   & 71.3     & 67.3 & 79.5 & \textbf{92.5}  & \textbf{85.1} & 74.2 & 54.8 & \textbf{63.6} & 74.4  & \textbf{48.2} & \textbf{64.8} & \textbf{74.1} & 61.6 & 38.3  & 71.1  & \textbf{100.0} & \textbf{82.8} & 86.5 & \textbf{65.4}  \\ 
                      & 3D-MPA \cite{engelmann20203d}      & \textbf{72.4}     & \textbf{69.9} & \textbf{83.4} & 87.6  & 76.1 &\textbf{74.8} & \textbf{56.6} & 62.2 & \textbf{78.3}  & 48.0 & 62.5 & 69.2 & \textbf{66.0} & \textbf{61.4}  & \textbf{93.1}  & 99.2  & 75.2 & \textbf{90.3} & 48.6  \\ \hline
\multirow{7}{*}{Weak} & DSRG         & 35.1     & 18.8  & 42.9 & 66.9  & 57.6 & 18.8 & 31.6  & 19.9  & 24.9   & 2.9  & 4.1  & 2.3 & 28.8  & 42.2   & 67.5  & 83.1  & 24.9  & 75.3 & 23.7   \\
                      & BoxMask      & 58.4     & 43.3 & \textbf{88.3} & 88.1  & 86.2 & 65.4 & 48.3 & 42.4  & 46.3  & 7.9  & 39.8 & 50.1 & \textbf{48.7}  & 48.2  & 74.6  & 95.6  & 56.0 & 82.9 & 44.6   \\
                      & DetIns       & 56.9     & 39.2  & 85.6 & 86.5  & 85.6 & 60.8 & 46.7 & 38.3  & 44.6  & 8.6 & 35.6 & 37.3 & 43.3 & \textbf{52.9}  & 78.6  & 94.6  & \textbf{64.2} & 78.2  & 40.9   \\ \cline{2-21} 
                      & SP           & 57.2     & 43.2 & 84.9 & 87.5  & 86.6 & 61.3 & 45.9 & 39.9 & 45.7  & 8.1  & 40.8 & 42.9  & 39.5 & 42.9  & 76.1  & 95.7  & 57.3 & 83.9 & 45.3  \\
                      & PC           & 57.5     & 42.6 & 87.5 & 86.7  & 83.8 & 62.9 & 48.5 & 37.9 & 45.7  & 7.2  & 41.6  & 46.6  & 43.6 & 44.3  & 78.6  & 94.7  & 63.2 & 84.3 & 41.2  \\
                      & SP + PC      & 59.1     & 43.4 & 84.9 & 88.4  & 86.6 & 66.2 & 47.0 & 40.0 & 45.9  & 7.3  & 46.0  & \textbf{55.5}  & 38.7 & 47.7  & 77.3  & 94.2  & 60.6 & \textbf{87.9} & 44.5  \\
                      & SP + PC + OR & \textbf{61.4}     & \textbf{45.4} & 86.8 & \textbf{89.0}  & \textbf{88.1} & \textbf{66.2} & \textbf{49.2} & \textbf{41.9} & \textbf{48.8}  & \textbf{9.9}  & \textbf{49.6}  & 52.3  & 47.8 & 48.3  & \textbf{82.6}  & \textbf{95.9}  & 63.2 & 87.4 & \textbf{45.5}  \\ \hline \hline
\end{tabular}}}
\end{table*}

\subsection{Comparison with Fully-supervised Methods}
We first show the effectiveness of our SPIB architecture under different ratios of labeled data from the entire training set of ScanNet v2 dataset. To be specific, we train Votenet only with the labeled data and regard the points within each bounding box as one instance, which are then evaluated under the setting of instance segmentation. SPCR is trained both with the labeled data and unlabeled data in a semi-supervised manner, and its performance of instance segmentation is evaluated in the same way as VoteNet. We conduct the comparison experiments against VoteNet and SPCR on the unseen validation set as listed in Table~\ref{semi_weak}. It is shown that our SPIB outperforms VoteNet and SPCR under all the label ratio settings for different evaluation metrics. Given 10$\%$ labeled data, our network obtains 9.4$\%$, and 12.5$\%$ absolute improvements for mAP@50$\%$ and mAP@25$\%$ respectively compared with VoteNet, which demonstrates that our SPIB is able to learn from unlabeled data especially when the labeled data is rare. Additionally, SPIB can achieve comparable performance with VoteNet when the label ratio is set as 40$\%$. Moreover, we are also able to further improve the performance with 100$\%$ labeled data. They indicate that our proposed weakly- and semi-supervised architecture is indeed effective.

 We further conduct the performance comparison between SPIB and the recent fully-supervised methods on ScanNet v2 dataset, by using all the training samples. To the best of our knowledge, our method is one of the pioneer works for weakly- and semi-supervised point cloud instance segmentation. Specifically, our method only utilizes bounding box labels for learning and predicting dense point cloud labels. Hence, we compare our method with the recent state-of-the-art fully supervised methods such as PointGroup \cite{Jiang_2020_CVPR} and 3D-MPA \cite{engelmann20203d}, to demonstrate the effectiveness of our weakly- and semi-supervised method using partial bounding box labels. As shown in Table~\ref{full_25} and Table~\ref{full_50}, one can observe that our network using bounding boxes as supervision can accomplish 61.4$\%$ on the mAP@25$\%$ scores, outperforming several recent fully supervised methods such as MRCNN, SGPN, and 3D-SIS. Especially, the mAP@25$\%$ scores are 39.2$\%$ higher than SGPN. However, we are also behind some other methods such as PointGroup and 3D-MPA due to much weaker and coarser labels in our setting. 

In some classes such as sofa and bed, our method can achieve 88.1$\%$ and 86.8$\%$ of mAP@25$\%$ respectively. These categories always have a regular geometry structure, which is a favorable factor for semantic propagation. Besides, the occupancy ratio of these categories always keeps the same for each object, so occupancy ratio guided refinement could play a bigger role among these categories. However, the performance of some classes such as picture and counter is bad with only 9.9$\%$ and 47.8$\%$ of mAP@25$\%$ scores. This is due to that the bounding box proposals of these categories are always flat and even similar to a plane, which is a big trouble for seed region expansion and property smoothing.

\begin{figure*}[ht]
    \centering
    \includegraphics[scale=0.39]{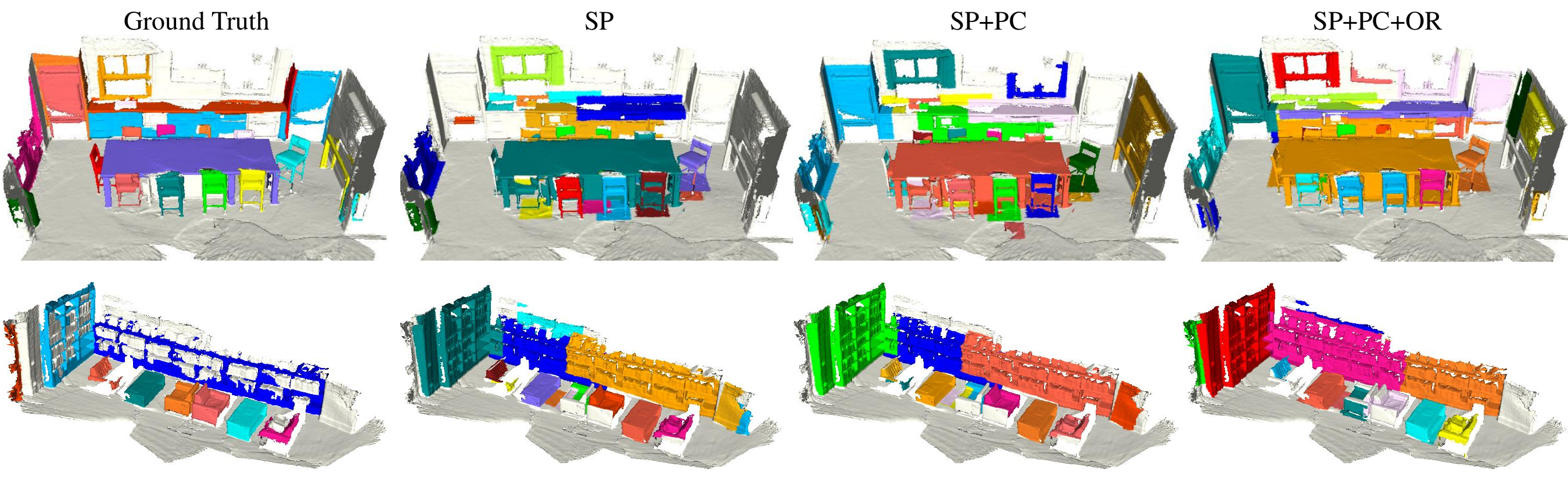}
	\caption{\label{visual_module}Qualitative results on ScanNet v2 dataset. SP represents semantic propagation component, PC represents property consistency graph, and OR represents occupancy ratio guided refinement. Different colors represent separate instances. The color of ground truth and predicted instances are not necessary to be the same.}
\end{figure*}

\begin{table*}[]
\caption{\label{full_50} Comparison with fully supervised methods. Per class 3D point cloud instance segmentation on ScanNet v2 test set with mAP@25$\%$ on 18 classes. All the methods are fully supervised except Ours.}
\renewcommand\arraystretch{1.4}
\setlength{\tabcolsep}{0.9mm}{
\scalebox{1.0}{
\begin{tabular}{c||c|ccccccccccccccccccc}
\hline
\hline
Mode &Method     & mAP@25$\%$ & bath  & bed   & bkshf & cab   & chair & cntr  & curt  & desk  & door  & ofurn & pic   & fridg & showr & sink  & sofa  & tabl  & toil  & wind  \\ \hline
\multirow{11}{*}{Full} & MRCNN \cite{he2017mask} & 26.1    & 90.3 & 8.1 & 0.8& 23.3 & 17.5 & 28.0 & 10.6 & 15.0 & 20.3 & 17.5 & 48.0 & 21.8 & 14.3 & 54.2 & 40.4 & 15.3 & 39.3 & 4.9 \\
&SGPN \cite{wang2018sgpn}       & 39.0    & 55.6 & 63.6 & 49.3 & 35.3 & 53.9 & 27.1 & 16.0 & 45.0 & 35.9 & 17.8 & 14.6 & 25.0 & 14.3 & 34.7 & 69.8 & 43.6 & 66.7 & 33.1 \\
                      & 3D-BEVIS \cite{elich20193d}   & 40.1    & 66.7 & 68.7 & 41.9 & 13.7 & 58.7 & 18.8 & 23.5 & 35.9 & 21.1 & 9.3 & 8.0 & 31.1 & 57.1 & 38.2 & 75.4 & 30.0 & 87.4 & 35.7 \\
                      &HCFS3D \cite{tan2021hcfs3d}       & 54.0    & 100.0 & 72.7 & 62.6 & 46.7 & 69.3 & 20.0 & 41.2 & 48.0 & 52.8 & 31.8 & 7.7 & 60.0 & 68.8 & 38.2 & 76.8 & 47.2 & 94.1 & 35.0 \\
                      & GSPN \cite{yi2019gspn}       & 54.4    & 50.0 & 65.5 & 66.1 & 66.3 & 76.5 & 43.2 & 21.4 & 61.2 & 58.4 & 49.9 & 20.4 & 28.6 & 42.9 & 65.5 & 65.0 & 53.9 & 95.0 & 49.9\\
                      &3D-SIS \cite{hou20193d}     & 55.8   & 100.0 & 77.3& 61.4& 50.3& 69.1& 20.0& 41.2& 49.8& 54.6& 31.1& 10.3& 60.0& 85.7& 38.2& 79.9& 44.5& 93.8& 37.1\\
                      &MASC \cite{liu2019masc}     & 61.5   & 71.1 & 80.2& 54.0& 75.7& 77.7& 2.9& 57.7& 58.8& 52.1& 60.0& 43.6& 53.4& 69.7& 61.6& 83.8& 52.6& 98.0& 53.4\\
                      &PE \cite{zhang2021point} & 77.6   & 100.0 & 90.0& 86.0& 72.8& \textbf{86.9}& 40.0& 85.7& 77.4& 56.8& 70.1& 60.2& 64.6& 93.3 &84.3& \textbf{89.0}& 69.1& 99.7& 70.9\\
                      &PointGroup \cite{Jiang_2020_CVPR} & 77.8   & 100.0 & 90.0& 79.8& 71.5& 86.3& 49.3& 70.6& \textbf{89.5}& 56.9& \textbf{70.1}& 57.6& 63.9& 100.0 &88.0& 85.1& 71.9& 99.7& 70.9\\
                      &SSTNet \cite{liang2021instance} & 78.9   & 100.0 & 84.0& \textbf{88.8}& 71.7& 83.5& \textbf{71.7}& 68.4& 62.7&\textbf{ 72.4}& 65.2& \textbf{72.7}& 60.0& 100.0 &91.2& 82.2& 75.7& \textbf{100.0}& 69.1\\
                      &HAIS \cite{chen2021hierarchical} & \textbf{80.3}   & 100.0 &\textbf{99.4}& 82.0& \textbf{75.9}& 85.5& 55.4& \textbf{88.2}& 82.7& 61.5& 67.6& 63.8& \textbf{64.6}& \textbf{100.0} &\textbf{91.2}& 79.7& \textbf{76.7}& 99.4& \textbf{72.6}\\ \hline
Weak                & SPIB (Ours)       & 63.4   & \textbf{100.0} & 86.4& 76.7& 58.5& 84.8& 53.5& 40.7& 49.2& 63.6& 50.8& 17.8& 49.5& 77.7& 49.2& 86.3& 52.6& 91.7& 51.7\\ \hline
\hline
\end{tabular}
}}
\end{table*}

\subsection{Ablation Studies}
We conduct five types of ablation studies, including the effect of SPCR, the baseline studies, the analysis of the contribution of each proposed component, the effects of different properties in property consistency graph, and the occupancy ratio prediction.

\begin{table*}[]
\caption{\label{semi_det} Comparison of semi-supervised learning with VoteNet on ScanNet v2 val set. Absolute improvements between our SPCR and VoteNet are reported in terms of mAP@50$\%$ and mAP@25$\%$ respectively.}
\renewcommand\arraystretch{1.4}
\setlength{\tabcolsep}{4.0mm}{
\scalebox{1.0}{
\begin{tabular}{c||c|cccccccccc}
\hline
\hline
Method                        & Metric   & 10\%         & 20\%         & 30\%         & 40\%         & 50\%         & 60\%         & 70\%         & 80\%         & 90\%         & 100\%        \\ \hline
\multirow{2}{*}{VoteNet \cite{qi2019deep}}      & mAP@50\% & 11.9         & 19.3         & 23.2         & 26.6         & 28.8         & 31.8         & 31.0         & 34.1         & 34.5         & 34.5         \\
                              & mAP@25\% & 29.6         & 41.2         & 46.3         & 47.9         & 51.6         & 54.1         & 54.9         & 55.0         & 57.6         & 58.1         \\ \hline
\multirow{2}{*}{SPCR}         & mAP@50\% & 18.0         & 25.6         & 30.3         & 33.2         & 33.8         & 35.4         & 36.6         & 36.4         & 37.8         & 37.4         \\
                              & mAP@25\% & 38.8         & 48.3         & 52.1         & 55.0         & 55.7         & 56.3         & 58.3         & 58.4         & 59.4         & 60.2         \\ \hline
\multirow{2}{*}{Improvements} & mAP@50\% & 6.1          & 6.3          & \textbf{7.1} & 6.6          & \textbf{5.0} & \textbf{3.6} & \textbf{5.6} & 2.3          & \textbf{3.3} & \textbf{2.9} \\
                              & mAP@25\% & \textbf{9.2} & \textbf{7.1} & 5.8          & \textbf{7.1} & 4.1          & 2.2          & 3.4          & \textbf{3.4} & 1.8          & 2.1          \\ \hline \hline
\end{tabular}}}
\end{table*}

\textbf{Effects of SPCR.} To study the effect of our semi-supervised bounding box proposal generation network SPCR, we compare SPCR with VoteNet on different label ratios and the results are reported in Table \ref{semi_det}. It can be seen that our SPCR works better than VoteNet under all the label ratio settings for both mAP@50\% and mAP@25\%. Moreover, we apply our SPCR on the newly data-efficient detection benchmark of ScanNet and achieve better performance than the recent state-of-the-art method CSC \cite{hou2020exploring} as listed in Table \ref{nb}, which demonstrates that our SPCR architecture is indeed effective for semi-supervised bounding box proposal generation.

\begin{table}[]
\caption{\label{nb} Comparison on the data-efficient detection benchmark of ScanNet. The metric is mAP@50\%}
\renewcommand\arraystretch{1.4}
\setlength{\tabcolsep}{4.2mm}{
\scalebox{1.0}{
\begin{tabular}{c||ccc}
\hline
\hline
Data Percentage & VoteNet \cite{qi2019deep} & CSC \cite{hou2020exploring}   & SPCR \\ \hline
10\%            & 0.3     & 8.6  & \textbf{18.0} \\
20\%            & 4.6     & 20.9 & \textbf{25.6} \\
40\%            & 22.0    & 29.2 & \textbf{33.2} \\
80\%            & 33.7    & \textbf{36.7} & 36.4 \\ \hline \hline
\end{tabular}}}
\end{table}

\begin{table}[]
\caption{\label{pc}Ablation study of property consistency term on the ScanNet v2 validation set.}
\renewcommand\arraystretch{1.4}
\setlength{\tabcolsep}{3.2mm}{
\scalebox{1.0}{
\begin{tabular}{ccc||cc}
\hline
\hline
coordinate & color & normal & mAP@50\% & mAP@25\% \\ \hline
          \Checkmark &      \XSolid &       \XSolid &32.6         &55.1          \\
          \XSolid &      \Checkmark &     \XSolid   &31.5        & 54.3         \\
          \XSolid &      \XSolid &       \Checkmark &31.2         &55.2          \\
          \Checkmark &     \Checkmark  &      \XSolid  &33.5         &56.1          \\
          \Checkmark &      \XSolid &        \Checkmark&32.8         &56.3          \\
          \XSolid &       \Checkmark&      \Checkmark  &32.9         &55.9          \\
          \Checkmark &       \Checkmark&       \Checkmark &\textbf{34.1}         &\textbf{57.5}          \\ \hline
          \hline
\end{tabular}}}
\end{table}

\begin{figure}[]
    \centering
    \includegraphics[scale=0.4]{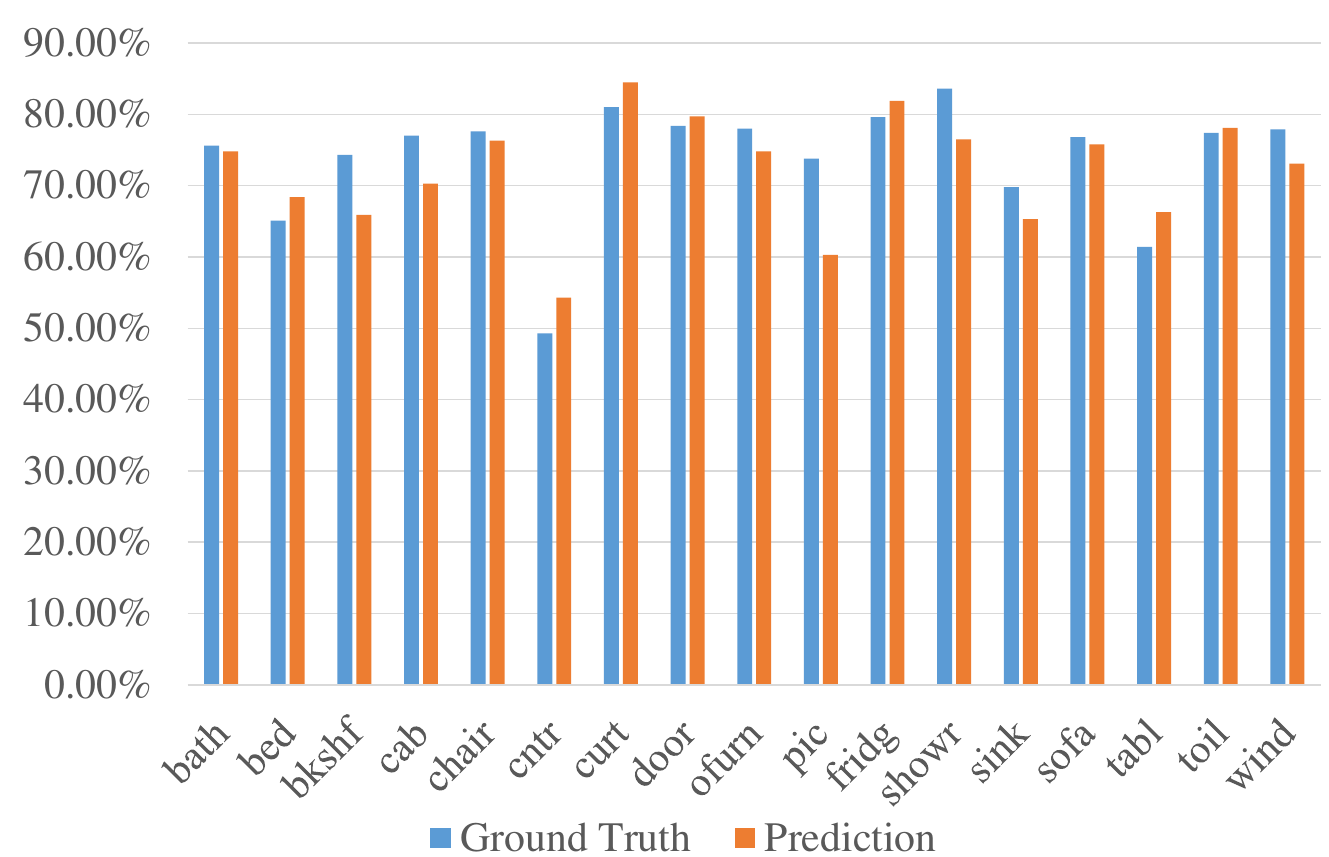}
	\caption{\label{occu_ratio}Occupancy ratio using ground truth bounding boxes vs. the predicted bounding boxes.}
\end{figure}

\textbf{Baseline Studies.} We develop three baseline methods for investigating how well the proposed method utilizes box-level annotations. The first baseline, DSRG, extends DSRG~\cite{huang2018weakly} algorithm in 2D image for point cloud instance segmentation leveraging proposals generated by VoteNet. In the second baseline, BoxMask first regards all the points within each ground truth bounding box as coarse instance annotations. Then, we add a segmentation head on top of VoteNet for instance segmentation as inspired by Mask-RCNN in 2D. Finally, we train the BoxMask baseline  using the above network and coarse instance annotations. The third baseline, DetIns first performs point cloud object detection using VoteNet. After that, we regard all the points within each predicted bounding box as one instance segmentation mask, and evaluates its performance under the setting of instance segmentation. Table~\ref{full_25} demonstrate the performance of the three baseline methods on ScanNet. One can see that their performance is worse than our proposed method, showing that our method utilizes the box-level supervisions much more effectively.

\textbf{Component Contributions.} We analyze the contributions of each proposed component of our network, including semantic propagation component, property consistency graph component, and occupancy ratio guided refinement component. The results are reported in Table~\ref{full_25}. We add our proposed components one by one to the baseline network DetIns. For the semantic propagation component, the improvements are gained in mAP@25$\%$ of 0.9$\%$ on ScanNet v2. The combination of semantic propagation component and property consistency graph further increases the former scores to 59.1$\%$. At last, applying all the three sub-components leads to the highest increases in mAP@25$\%$ of 4.5$\%$, which demonstrates that all components are effective. Moreover, we have calculated the average time for the three different modules for processing one sample to represent their running time. Specifically, the average running time of SP, PC, and OR are 0.048 seconds, 0.038 seconds, and 0.052 seconds respectively. Qualitative results on the ScanNet v2 dataset are visualized in Fig.~\ref{visual_module}.

\textbf{Effects of Property Terms.} We further study the effects of the three property terms in our proposed property consistency graph with different combinations. The results are reported in Table~\ref{pc}. For three individual property terms, the coordinate and color property terms have more contributions than the normal term, and the coordinate term performs the best alone. Besides, the combination of each two terms produces better performance than their terms solely. Finally, the integration of all the three property terms outperforms any other combination, which proves that properties of points on each instance indeed boost the process of instance mask prediction.

\textbf{Occupancy Ratio Prediction.}
Fig.~\ref{occu_ratio} shows that the occupancy ratio of different classes varies from each other. The estimated occupancy ratios using predicted bounding boxes' masks are quite similar to the ratio predicted using ground-truth boxes' masks, which demonstrates the effectiveness of our occupancy ratio learning.

\section{Conclusion}
In this work, we propose the first semi-supervised point cloud instance segmentation network with bounding boxes as supervision. We propose a novel semantic propagation component and a property consistency graph, to mine the instance masks within predicted bounding boxes on the learned semantic score map and original point cloud respectively. Moreover, we observe that the occupancy ratio of each instance is related to its semantic class, which inspires us to calculate the occupancy ratio of each category to refine the instance candidates predicted by the above two components. Our method achieves comparable or even better performance when compared with recent fully supervised methods on the challenging ScanNet v2 dataset, thus demonstrates that our weakly- and semi-supervised setting is a promising learning paradigm to solve the 3D point cloud scene perception problem.

\ifCLASSOPTIONcompsoc
  \section*{Acknowledgments}
\else
  \section*{Acknowledgment}
\fi

This work is supported by National Natural Science Foundation of China (No. 62071127 and U1909207), Shanghai Pujiang Program (No.19PJ1402000), Shanghai Municipal Science and Technology Major Project (No.2021SHZDZX0103), Shanghai Engineering Research Center of AI Robotics and Engineering Research Center of AI Robotics, Ministry of Education in China and the Agency for Science, Technology and Research (A*STAR) under its AME Programmatic Funding Scheme (Project A18A2b0046).



%

%
%
%
%
%
%
%




\bibliographystyle{IEEEtran}
\bibliography{trans.bib}

\ifCLASSOPTIONcaptionsoff
  \newpage
\fi



%

%
%

%

\vspace{-10pt}
\begin{IEEEbiography}[{\includegraphics[width=1in,height=1.25in,clip,keepaspectratio]{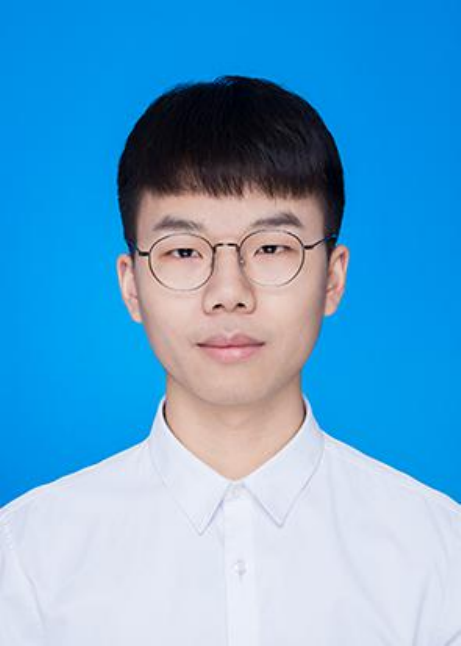}}]{Yongbin Liao}
received the bachelor’s degree in communication engineering from Wuhan University of Technology in July 2019. He is currently pursuing the master degree in School of Information Science and Technology, Fudan University. His research interests is weakly-supervised learning, semi-supervised learning and point cloud instance segmentation.
\end{IEEEbiography}
\vspace{-10pt}

\vspace{-10pt}
\begin{IEEEbiography}[{\includegraphics[width=1in,height=1.25in,clip,keepaspectratio]{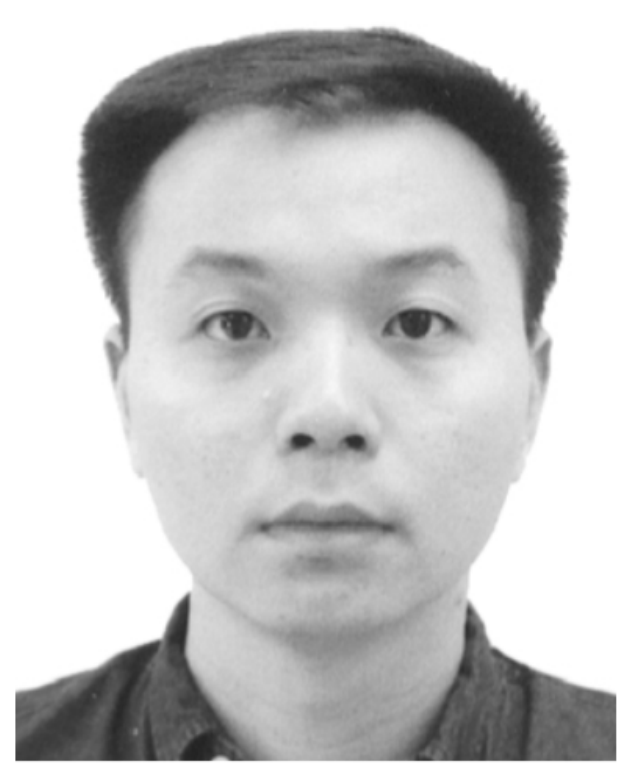}}]{Hongyuan Zhu}
received the B.S.degree in software engineering from the University of Macau, Macau, in 2010, and the Ph.D. degree in computer engineering from Nanyang Technological University, Singapore, in 2014. He is currently a Research Scientist with the Institute for Infocomm Research, A*STAR, Singapore. His research interests include multimedia content analysis and segmentation.
\end{IEEEbiography}
\vspace{-10pt}

\vspace{-10pt}
\begin{IEEEbiography}[{\includegraphics[width=1in,height=1.25in,clip,keepaspectratio]{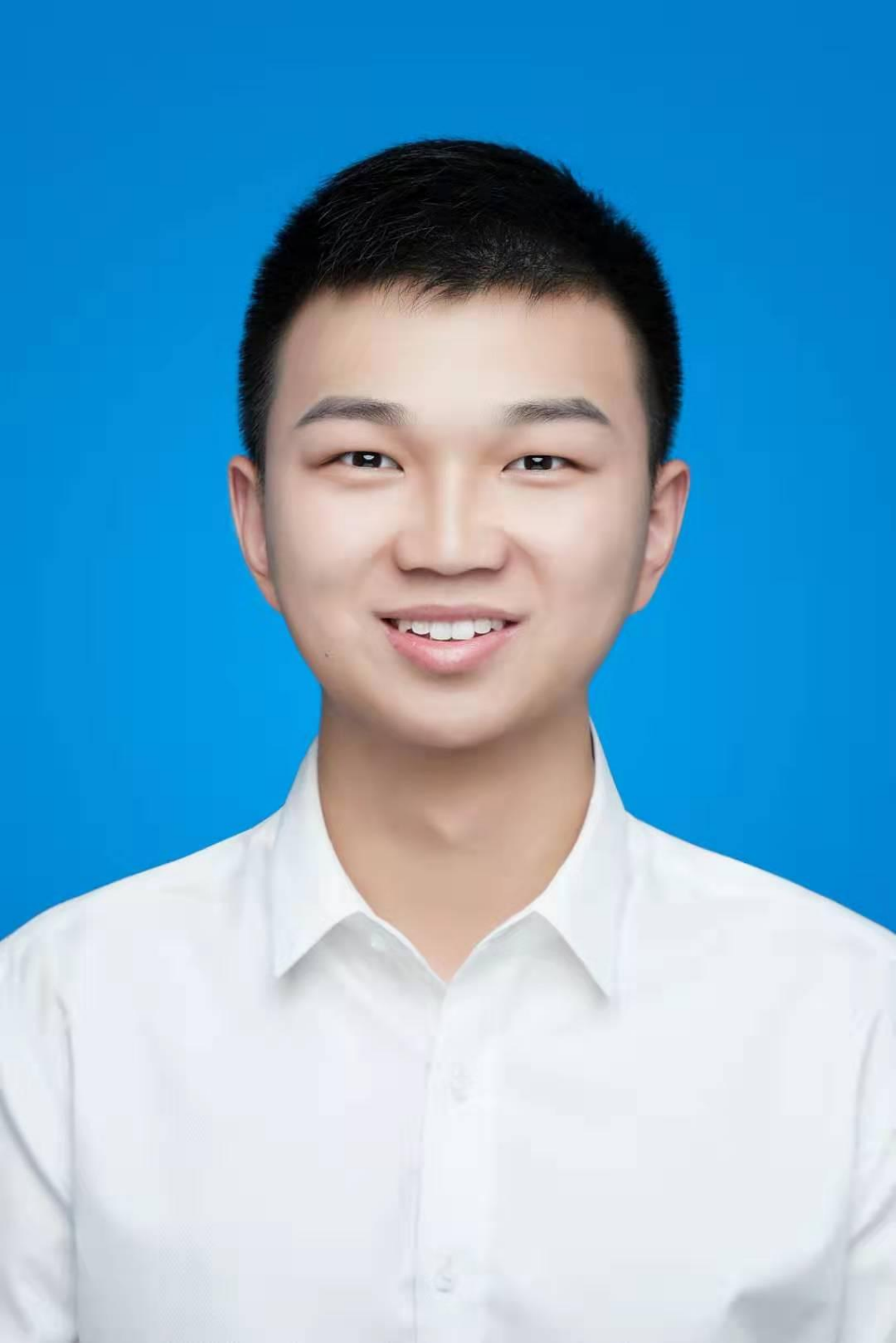}}]{Yanggang Zhang}
is studying as a master student in School of Information Science and Technology, Fudan University. He graduated from Nanjing Forestry University in 2019 with bachelor degree, now his research direction is deep learning algorithms in 3D object detection and semantic segmentation.
\end{IEEEbiography}
\vspace{-10pt}

\vspace{-10pt}
\begin{IEEEbiography}[{\includegraphics[width=1in,height=1.25in,clip,keepaspectratio]{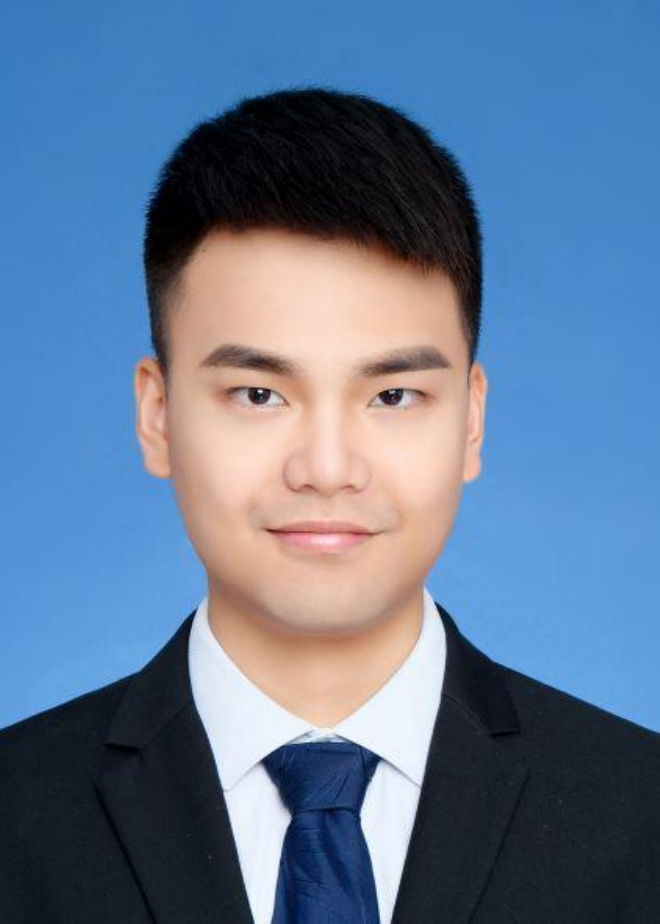}}]{Chuangguan Ye}
is studying as a master student in School of Information Science and Technology, Fudan University. His research interests is few-shot learning and point cloud classification.
\end{IEEEbiography}

\vspace{-10pt}
\begin{IEEEbiography}[{\includegraphics[width=1in,height=1.25in,clip,keepaspectratio]{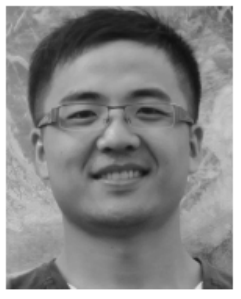}}]{Tao Chen}
received the Ph.D. degree in information engineering from Nanyang Technological University, Singapore, in 2013. He was a Research Scientist at the Institute for Infocomm Research, A*STAR, Singapore, from 2013 to 2017, and a Senior Scientist at
the Huawei Singapore Research Center from 2017 to 2018. He is currently a Professor with the School of Information Science and Technology, Fudan University, Shanghai, China. His main research interests include computer vision and machine learning.
\end{IEEEbiography}
\vspace{-10pt}

\vspace{-10pt}
\begin{IEEEbiography}[{\includegraphics[width=1in,height=1.25in,clip,keepaspectratio]{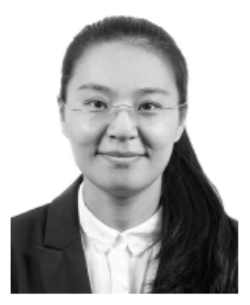}}]{Jiayuan Fan}
received the Ph.D. degree in information engineering from Nanyang Technological University, Singapore, in 2015. She was a Research Scientist at the Institute for Infocomm Research, A*STAR, Singapore, from 2015 to 2018. She is currently a Professor with the Academy for Engineering and Technology, Fudan University, Shanghai, China. Her main research interests include computer vision, and image forensic analysis and application.
\end{IEEEbiography}




\end{document}